\documentclass[letterpaper, 10 pt, conference]{ieeeconf}  
\IEEEoverridecommandlockouts                            

\usepackage{graphics} 
\usepackage{epsfig} 
\usepackage{graphicx}
\usepackage{xcolor} 
\usepackage{capt-of}

\usepackage{amsmath} 
\usepackage{amssymb}  
\usepackage{amsfonts}
\usepackage{mathtools}
\usepackage{units}

\usepackage{booktabs}
\usepackage{multirow}
\usepackage{tabularx}
\usepackage{multirow}

\usepackage{hyperref}
\usepackage{balance}
\usepackage{cite}

\usepackage{tikz}
\usepackage{pgfplots}
\usepackage{pgfplotstable}
\usepackage{circuitikz}
\usetikzlibrary{bending, plotmarks, shapes, pgfplots.statistics, pgfplots.colorbrewer, positioning}

\pgfplotsset{scaled y ticks=false}
\pgfplotsset{compat=1.16,
    width = 7cm,
    height = 3.5cm,
    title style = {yshift=-6pt},
    xlabel shift = -3pt,
    ylabel shift = -0pt,
    cycle list={{matlab1},{matlab2},{matlab3},{matlab9},{matlab4},{matlab5},{matlab6},{matlab7},{matlab8}},
    legend columns = 1,
    legend cell align={left},
    legend style ={
        draw = gray,
        fill opacity=0.8,
        text opacity=1.0,
        draw opacity=1.0,
    },
    xmajorgrids,
    ymajorgrids,
    scale only axis,
} 
\usepackage{xfrac}

\usepackage[T1]{fontenc}

\newcommand{\rotateRPY}[3]
{   \pgfmathsetmacro{\rollangle}{#1}
    \pgfmathsetmacro{\pitchangle}{#2}
    \pgfmathsetmacro{\yawangle}{#3}

    \pgfmathsetmacro{\newxx}{cos(\yawangle)*cos(\pitchangle)}
    \pgfmathsetmacro{\newxy}{sin(\yawangle)*cos(\pitchangle)}
    \pgfmathsetmacro{\newxz}{-sin(\pitchangle)}
    \path (\newxx,\newxy,\newxz);
    \pgfgetlastxy{\nxx}{\nxy};

    \pgfmathsetmacro{\newyx}{cos(\yawangle)*sin(\pitchangle)*sin(\rollangle)-sin(\yawangle)*cos(\rollangle)}
    \pgfmathsetmacro{\newyy}{sin(\yawangle)*sin(\pitchangle)*sin(\rollangle)+ cos(\yawangle)*cos(\rollangle)}
    \pgfmathsetmacro{\newyz}{cos(\pitchangle)*sin(\rollangle)}
    \path (\newyx,\newyy,\newyz);
    \pgfgetlastxy{\nyx}{\nyy};

    \pgfmathsetmacro{\newzx}{cos(\yawangle)*sin(\pitchangle)*cos(\rollangle)+ sin(\yawangle)*sin(\rollangle)}
    \pgfmathsetmacro{\newzy}{sin(\yawangle)*sin(\pitchangle)*cos(\rollangle)-cos(\yawangle)*sin(\rollangle)}
    \pgfmathsetmacro{\newzz}{cos(\pitchangle)*cos(\rollangle)}
    \path (\newzx,\newzy,\newzz);
    \pgfgetlastxy{\nzx}{\nzy};
}

\newcolumntype{C}{>{\centering\arraybackslash}X}
\newcolumntype{x}[1]{>{\centering\let\newline\\\arraybackslash\hspace{0pt}}p{#1}}
\definecolor{matlab1}{rgb}{0.00000,0.44700,0.74100}
\definecolor{matlab2}{rgb}{0.85000,0.32500,0.09800}
\definecolor{matlab3}{rgb}{0.92900,0.69400,0.12500}
\definecolor{matlab4}{rgb}{0.49400,0.18400,0.55600}
\definecolor{matlab5}{rgb}{0.4660, 0.6740, 0.1880}
\definecolor{matlab6}{rgb}{0.3010, 0.7450, 0.9330}
\definecolor{matlab7}{rgb}{0.6350, 0.0780, 0.1840}
\definecolor{matlab8}{rgb}{0.8, 0.8, 0}
\definecolor{matlab9}{rgb}{0.6, 0.6, 0.6}
\definecolor{verylightgray}{rgb}{0.98,0.98,0.98}

\pgfdeclarelayer{bg}

\makeatletter
\g@addto@macro\@maketitle{
    \setcounter{figure}{0}
    \vspace*{12pt}
    \centering
    \includegraphics[width=17.6cm]{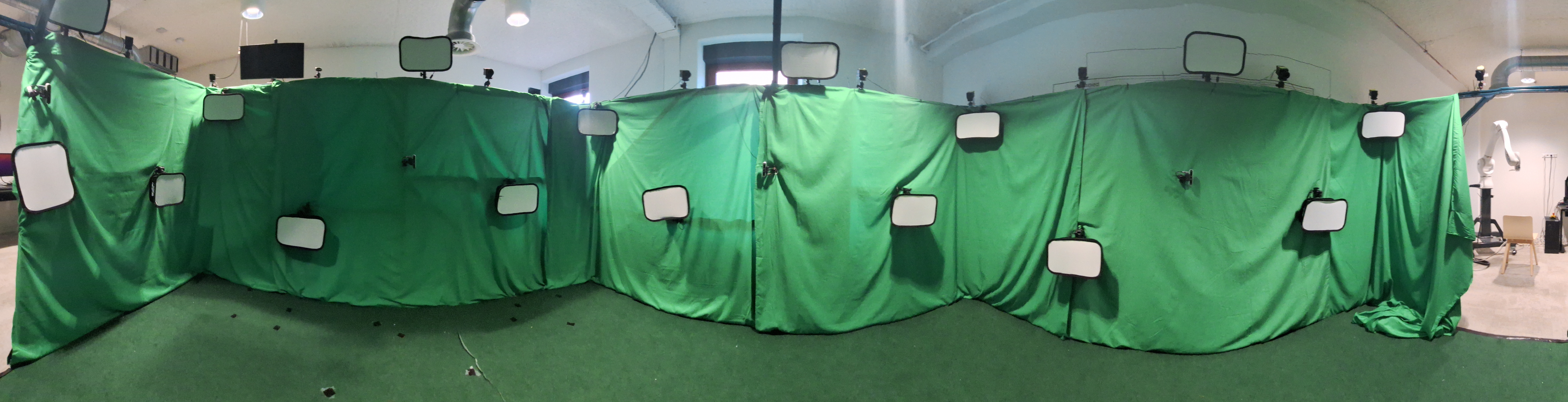}
    \vspace*{-3pt}
    \captionof{figure}{Panorama of the capture environment. Cameras and lights are mounted around the capture volume, while the green curtain and carpet create a uniform, controlled backdrop for consistent multi-view recording.}
    \label{fig:panorama}
    \vspace*{-12pt}
}

\title{\LARGE \bf
A Synchronized Audio-Visual Multi-View Capture System
}

\author{Xiangwei Shi, Gara Dorta, Ruud de Jong, Ojas Shirekar, Chirag Raman\\ 
\{x.shi-3, g.dorta, r.dejong, o.k.shirekar, c.a.raman\}@tudelft.nl \\
Delft University of Technology, Netherlands\\
}

\begin{document}

\makeatletter
\maketitle
\thispagestyle{empty}

\begin{abstract}
    Multi-view capture systems have been an important tool in research for recording human motion under controlling conditions. Most existing systems are specified around video streams and provide little or no support for audio acquisition and rigorous audio-video alignment, despite both being essential for studying conversational interaction where timing at the level of turn-taking, overlap, and prosody matters. In this technical report, we describe an audio-visual multi-view capture system that addresses this gap by treating synchronized audio and synchronized video as first-class signals. The system combines a multi-camera pipeline with multi-channel microphone recording under a unified timing architecture and provides a practical workflow for calibration, acquisition, and quality control that supports repeatable recordings at scale. We quantify synchronization performance in deployment and show that the resulting recordings are temporally consistent enough to support fine-grained analysis and data-driven modeling of conversation behavior.
\end{abstract}


\section{Introduction}
\label{sec:introduction}
Multi-camera capture studios are a widely used tool in computer vision and graphics research to record high-quality human motion and appearance to facilitate reconstruction, tracking, and analysis. Significant effort has been devoted to designing reliable multi-view video capture setups, addressing camera placement, calibration, synchronization, and data throughput to support downstream 3D reasoning from images. However, these research setups are often primarily specified around video. Audio capture is frequently unsupported, treated as secondary, or only briefly described, despite audio being central to the study of human communication and iteration.

This video-centric emphasis becomes a limiting factor for human interaction research, where conversational behavior is inherently multimodal. Phenomena such as turn-taking, backchannels, interruptions, and co-speech gestures depend on fine-grained timing between speech and motion. Misalignment between audio and video can obscure the temporal ordering that defines interactive dynamics. Moreover, interaction scenes pose visual challenges: multiple participants occlude each other and move unpredictably, making multi-view recording necessary for obtaining stable pose estimates and reliable 3D motion from 2D detections.

Recent advances in generative modeling further amplify these requirements. Tasks such as co-speech gesture generation~\cite{habibie2021learning,zhu2023taming,liu2022beat,qi2025co3gesture} and conversational motion synthesis~\cite{jin2019threepartyheadeye,jin2022s2mnet,canales2023realtimegaze} explicitly condition motion on speech signals and increasingly target interactive or multi-party settings where relative timing between participants is critical. Similarly, pose-controllable talking head generation~\cite{ma2023styletalk,liang2022expressive,zhou2021pose} treats head motion and body pose as key controllable factors alongside speech-driven mouth dynamics, again relying on consistent alignment between audio and visual cues. In these applications, a capture system that delivers high-quality multi-view video but lacks a robust, time-aligned audio stream risks producing data that is ill-suited for modeling, evaluation, or behavioral measurement.

A limited number of prior systems have treated synchronized audio as a first-class component of multi-view capture system. For example, MPI VideoLab~\cite{kleiner2004videolab} explicitly targets time-synchronized multi-view video and audio for recording human actions. More recently, industrial avatar pipelines~\cite{facebookFacebookBuilding} also describe capture procedures that require extensive synchronized video and audio recording in controlled lab environments, underscoring that realistic communication modeling depends on both modalities. These examples highlight a practical gap: while multi-view capture infrastructure is common, there is comparatively less "blueprint-level" documentation on how to build and operate a system where audio-video synchronization is engineered to the same standard as multi-camera video synchronization, specifically for interaction-centric data collection.

In this technical report, we present a synchronized audio-visual multi-view capture system designed around two coequal requirements: (i) dense multi-view visual coverage for robust observation of multi-person interaction, and (ii) reliable synchronization across cameras and between audio and video to support fine-grained analysis and multimodal learning. We describe the system architecture and operational workflow and report synchronization performance that enable audio-motion studies where precise timing is critical. Our goal is to provide a reproducible, practice-oriented reference for building capture infrastructure that supports modern audio-conditioned motion and talking head research, as well as broader studies of conversational behavior. 
\section{Related Work}
\label{sec:related_work}
\subsection{Foundations of multi-view video capture.} Multi-camera studios are a cornerstone technology for capturing dynamic scenes under controlled conditions, enabling tasks such as free-viewpoint video, 3D reconstruction, and markerless motion capture. Foundational “studio blueprint” systems~\cite{theobalt2003studio, matusik20043dtv, waschbusch2005scalable3dvideo} established the core principles, emphasizing camera synchronization, calibration, lighting, and high-throughput data pipelines. This line of work, continued and refined by studios such as those from the Max Planck Institute~\cite{theobalt2007high}, provides comprehensive guidance for building camera-centric infrastructure. However, these systems typically treat audio as out of scope or an ancillary concern. 

\subsection{Scaling to multi-person and social interaction.}
Capturing social interaction introduces additional challenges, including persistent occlusions, subtle motions, and the need for large volumes. Systems like the Panoptic Studio~\cite{joo2015panoptic} addressed these by scaling to hundreds of synchronized cameras, enabling robust 3D pose estimation in groups through dense multi-view evidence. While pushing the boundaries of visual capture for social settings, such systems remain primarily focused on video acquisition and 3D inference, with audio not integrated as a first-class, synchronized modality.

\subsection{Toward Integrated Audio-Visual Capture Systems.} Recognizing that many research questions in communication and behavior rely on precise multimodal timing, a subset of academic and industrial systems have begun to treat synchronized audio as a coequal design objective. In academia, the MPI VideoLab~\cite{kleiner2004videolab} exemplifies this approach, being explicitly designed for high-quality synchronous recording of video and audio from multiple viewpoints for psycholinguistic research. This principle is even more pronounced in industrial avatar and telepresence pipelines. For instance, Meta’s Codec Avatars~\cite{facebookFacebookBuilding} and the subsequent “Codec Avatar Studio”~\cite{martinez2024codec} rely on extensive, synchronized multi-view image and audio capture in specialized labs to produce high-fidelity models. These efforts~\cite{martinez2024codec, richard2021audiogaze} reflect a clear trend: advanced audio-conditioned and interaction-centric modeling necessitates capture infrastructure where audio-video synchronization is engineered with the same rigor as multi-camera synchronization.
\section{System Design}
\label{sec:system_design}
The capture system is built around a single dedicated studio, illustrated in Fig.~\ref{fig:panorama}, which integrates all necessary sensing, synchronization, and control infrastructure for recording human interaction and motion. The studio provides a controlled and reproducible environment, structured around a cubic green-screen frame that defines the capture volume and serves as the mounting structure for all equipment. Within this space, machine-vision cameras and a multi-channel audio array acquire synchronized video and audio under programmable lighting. A master timecode generator supplies a common timing reference to all devices, while the in-room control computers manage device configuration, recording, and initial data handling. The following subsections detail each component and its role in the end-to-end capture workflow.

\subsection{Capture volume frame.} The core structure within the capture system is a modular cubic capture frame, which defines the recording volume and acts as the primary mechanical mount for all equipments. Constructed from $45\times45$\textit{mm} aluminum extrusion for modularity and rigidity, the frame measures $3.5 \textit{m}\times3.5 \textit{m}\times2.0\textit{m}$ (height). This volume is sized to accommodate multi-person interactions -  up to approximately six participants - while maintaining sufficient camera coverage to mitigate severe occlusions. The frame is organized into three horizontal levels and features an access door on one side, allowing flexible placement of cameras and lighting at multiple heights. To suppress background appearance and simplify downstream processing, all four sides and the floor are covered with green curtains and a gree carpet, providing a consistent chroma-key backdrop. Fig.~\ref{fig:capture_frame} shows the structure of the capture frame.

\begin{figure}[t]
    \centering
    \includegraphics[width=0.23\textwidth]{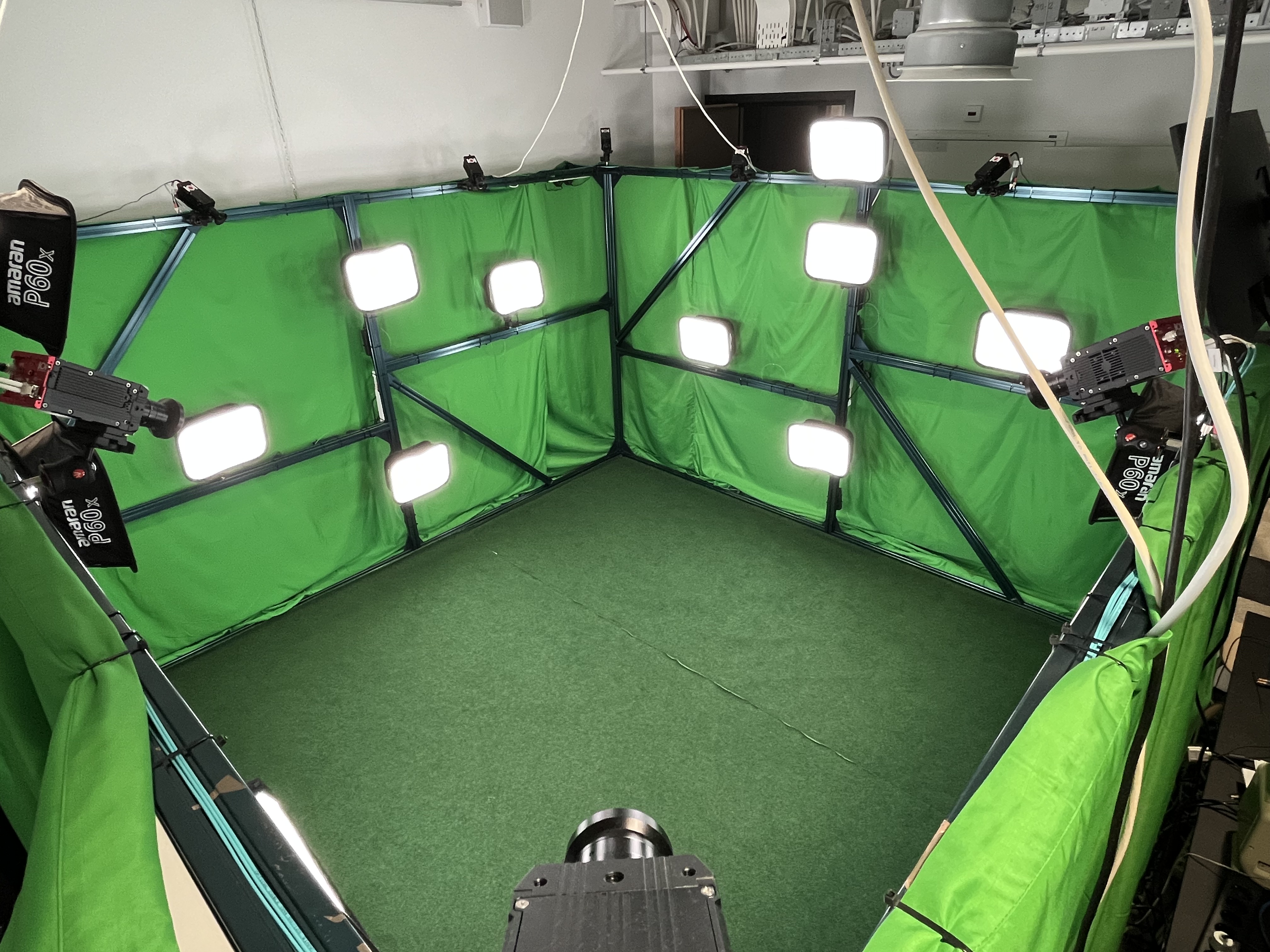}
    \hfill
    \includegraphics[width=0.23\textwidth]{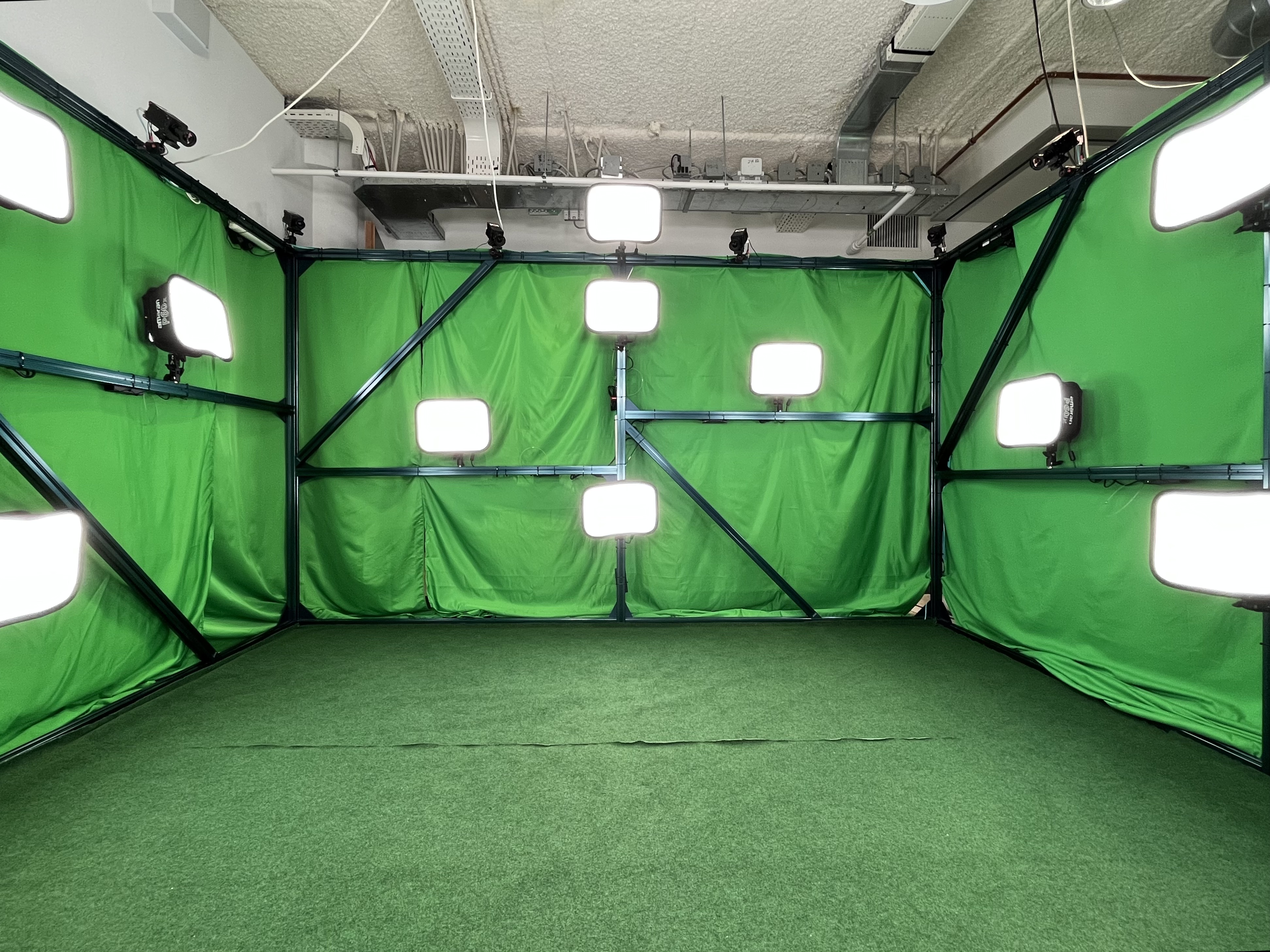}
    \caption{Top and interior views of the modular capture frame. The structure is organized into three horizontal mounting levels, allowing flexible placement of cameras and lighting at arbitrary positions.}
    \label{fig:capture_frame}
\end{figure}

\subsection{Cameras.} The studio is equipped with 12 \textsc{IO Industries} Volucam machine-vision cameras\footnote{https://www.ioindustries.com/cameras}, which employ global-shutter sensors to avoid rolling-shutter artifacts during fast human motion 
We use the 120B68 variant (\textsc{Sony} Pregius IMX253 sensor), capable of capturing at a resolution of $4096\times3008$ and up to 68 fps (depending on bit depth/recording mode). These cameras were selected primarily based on (i) high resolution, (ii) global shutter operation, (iii) hardware-level multi-camera synchronization support, and (iv) robust vendor-provided software for multi-camera control (StudioCap-VC client and an SDK for custom integration). Beyond these, the platform facilitates high-throughput recording through features such as internal RAW recording to SSD, a 10GbE interface for configuration and fast data transfer, and practical studio functionalities including optional SDI monitoring output and multiple synchronization inputs (PTP, LTC, GPIO), making it well-suited for deployment in large multi-camera arrays.

\begin{figure}[t]
    \centering
    \includegraphics[width=0.23\textwidth]{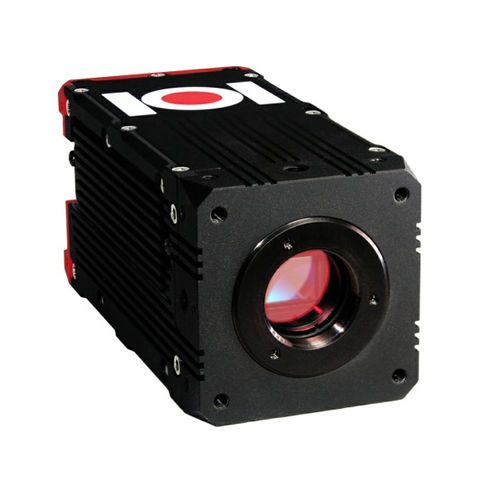}
    \hfill
    \includegraphics[width=0.23\textwidth]{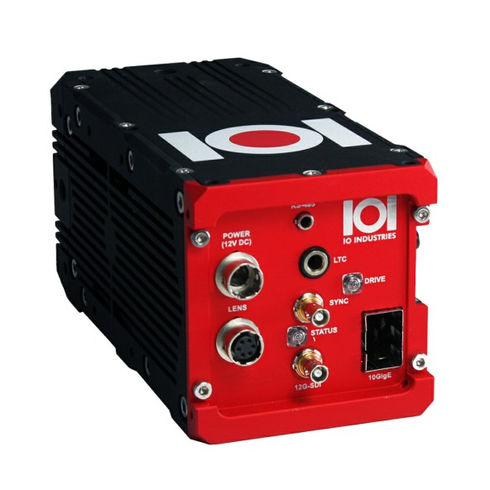}
    \caption{Overview of the camera unit. The right figure illustrates some key functionalities and supported synchronization inputs (e.g., LTC, SYNC).}
    \label{fig:capture_frame}
\end{figure}

\subsection{Lens.} Three types of lenses are employed to balance field of view and spatial detail according to the recording objective. For multi-person interaction capture, we primarily use \textsc{VST} fixed-focal C-mount lenses for wider coverage: the VS-0828HV (8\textit{mm}) and VS-1220HV (12\textit{mm}). For instance, the 8\textit{mm} lens provides a substaintially wider angle of view than the 12\textit{mm} lens. For higher-detail recordings, such as face-focused views, we use \textsc{Sigma} 18-35\textit{mm} f/1.8 DC HSM Art zoom lens, which offer a narrower field of view at longer focal lengths and greater flexibility in framing.

\subsection{Audio subsystem.} The audio subsystem is designed to capture clean, multi-speaker dialogue during social interaction while maintaining precise temporal alignment with the video streams. It consists of a scalable multi-channel wireless microphone setup and a profession multi-channel audio interface that supports both internal and external synchronization.

\subsection{Multi-channel wireless microphones.} The system is equipped with 32 channels of \textsc{Sennheiser} wireless microphones, each comprising a bodypack transmitter and a rack-mounted receiver. For lavalier capsules, we employ two microphone models: the omnidirectional \textsc{Sennheiser} MKE2 and the cardioid \textsc{Sennheiser} ME4.
In multi-person interaction recordings, the ME4 is predominantly used for its directional pickup pattern, which emphasizes the wearer's speech while attenuating ambient noise and the voices of nearby participants, which is a crucial feature when multiple speakers are in close proximity. The setup is operationally scalable: while typical sessions may utilize a subset of channels (e.g., six simultaneous speakers), the infrastructure can be expanded to accommodate more participants without modifying the room layout.

\subsection{Audio interface and conversion.} Multi-channel audio acquisition is managed by an \textsc{RME} Fireface UFX III audio interface, paired with an \textsc{RME} M-32 AD 32-channel analog-to-digital converter. Both units support synchronization via word clock, and the M-32 AD further enhances timing stability with \textsc{RME}'s SteadyClock technology for jitter suppression. This configuration allows the audio chain to operate from an internal reference or to be locked to an external studio clock based on session requirements. Controlled by a dedicated computer, the system records all audio channels simultaneously onto a unified timeline, establishing a reliable foundation for consistent audio-video alignment in subsequent processing.

\begin{figure}[t]
    \centering
    \includegraphics[width=0.295\textwidth]{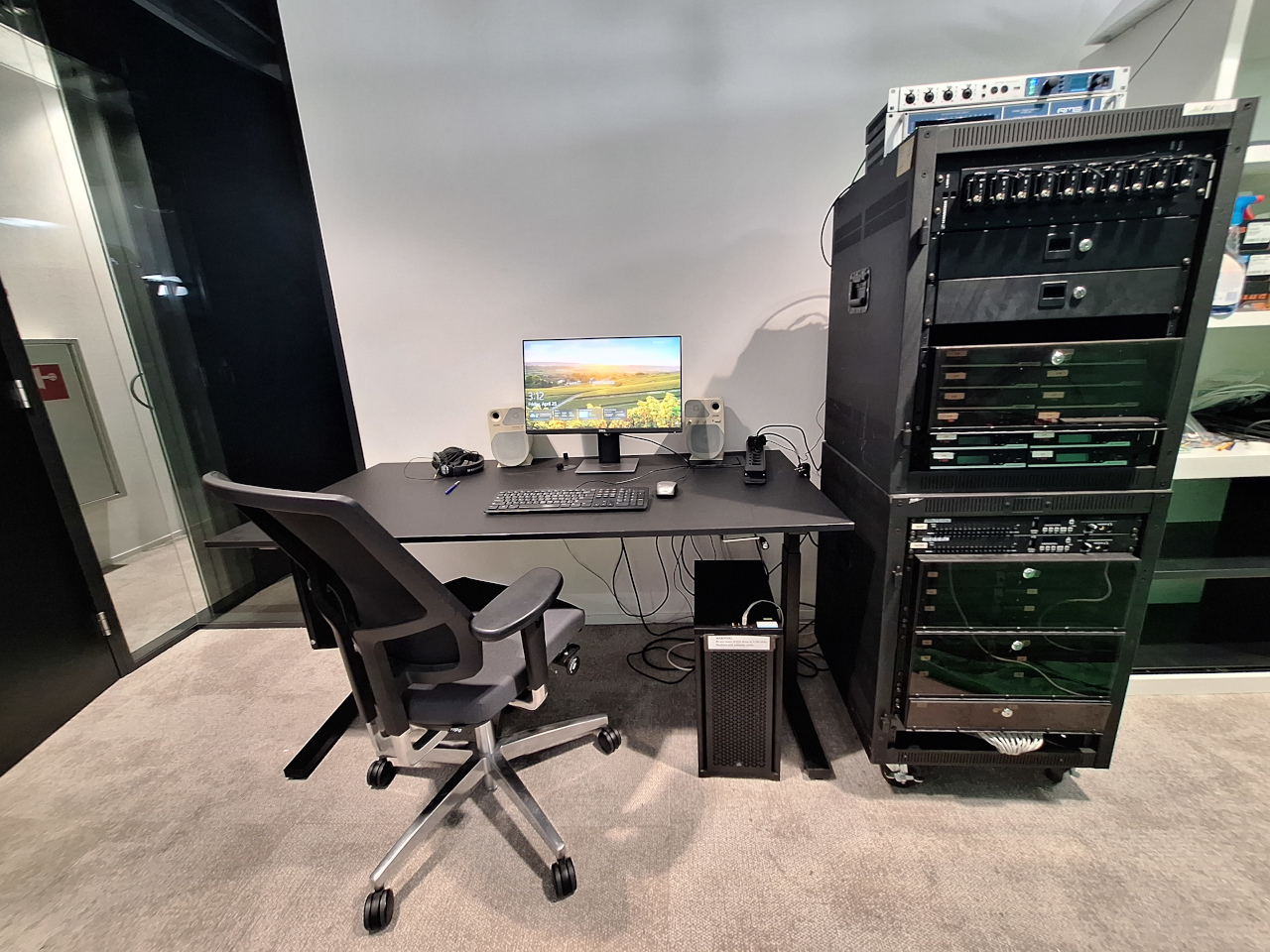}
    \hfill
    \includegraphics[width=0.18\textwidth]{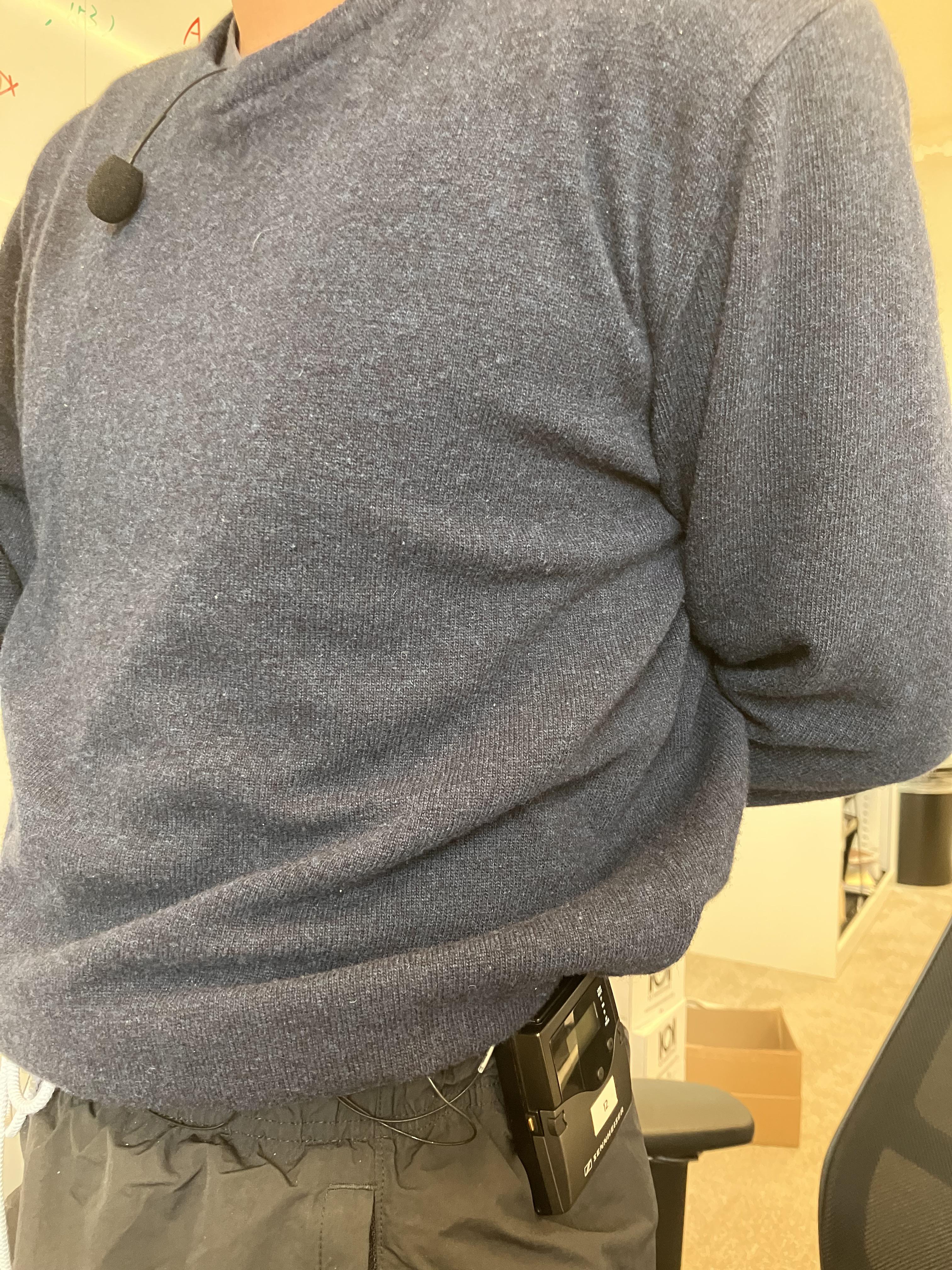}
    \caption{Left: the control computer and microphone rack housing the audio interface and wireless receivers. Right: a close-up of a wearable bodypack transmitter from a single microphone channel.}
    \label{fig:audio}
\end{figure}

\subsection{Network connectivity and storage.} All capture devices are integrated through a dedicated local 10GbE network that supports both device control and high-throughput data transfer. The 12 machine-vision cameras are connected to an \textsc{FS} S5850-series switch via SFP+ fiber links, ensuring each camera has a 10GbE connection and forming a compact, rack-mountable backbone suitable for multi-camera setups. The control computer, equipped with a 10GbE network interface, connects to the same switch via Ethernet, enabling centralized configuration and monitoring of all cameras while sustaining high aggregate data transfer rates. In contrast to the cameras, the audio chain does not rely on network transport. The control computer interfaces directly with the \textsc{RME} Fireface audio unit via USB, facilitating simultaneous multi-channel recording on the same host system. For storage, recordings are first written to local media on the control computer (e.g., external SSD) for reliable acquisition, and then transferred to a room-local NAS for consolidation and shared access. We use a \textsc{QNAP} TS-832PXU NAS with 8 drive bays and dual 10GbE SFP+ network interfaces connected to the same switch, providing sufficient bandwidth for rapid offload and multi-user access. With all eight bays populated, the system is configured as RAID5 to balance usable capacity and fault tolerance, yielding on the order of ~70 TB usable storage.

\begin{figure}[t]
    \centering
    \includegraphics[width=0.25\textwidth]{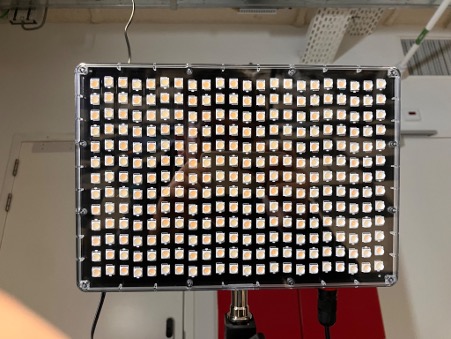}
    \hfill
    \includegraphics[width=0.20\textwidth]{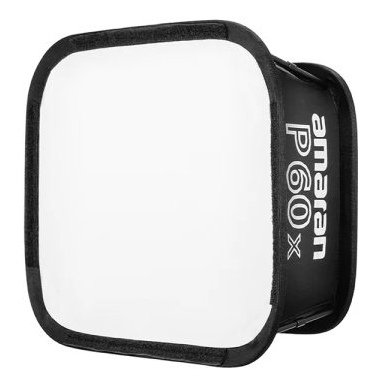}
    \caption{Left: LED panel. Right: light diffuser.}
    \label{fig:light}
\end{figure}

\subsection{Lighting.} To ensure consistent illumination across viewpoints and reduce motion blur in fast gestures and human motion, the capture volume is lit using 18 \textsc{amaran} P60x bi-color LED panels. The P60x lights support adjustable correlated color temperature from 3200\textit{K} to 6500\textit{K}, allowing the room lighting to be matched to different conditions as needed. In typical recordings we operate the panels at a fixed "daylight-like" setting, commonly around 5500\textit{K} and at high output too maintain stable exposure across all cameras. To improve visual comfort and image quality, each panel is used with a diffuser to soften shadows and reduce specular highlights on faces and skin.
\section{Synchronization}
\label{Synchronization}

Synchronization remains one of the most critical challenges in multimodal data acquisition, particularly for analyzing multiparty social dynamics where temporal relationships between modalities are essential for interpreting higher-order behavioral constructs. The precision with which different data streams are aligned directly impacts the validity of analyses that examine relationships across modalities, such as correlations between speech patterns and gestural movements or physiological responses to social stimuli.

We adopt a synchronization-at-acquisition approach based on the modular framework of Raman et al. ~\cite{raman2020modular}, previously deployed in the Conflab dataset for in-the-wild data collection in highly dynamic social settings~\cite{raman2022conflab}.
Specifically, we address synchronization at both intra- and inter-modal levels.

First, multi-camera synchronization establishes a unified temporal reference across all viewpoints. This is a prerequisite for any robust 3D reconstruction or motion analysis, as it ensures that images captured from different angles represent the same instantaneous state of the scene. Without this tight inter-camera alignment, triangulation fails, motion trajectories become blurred, and the resulting 3D models lose coherence. Second, audio-video synchronization is of equal importance for interaction studies. Human communication is inherently multimodal, where meaning is often encoded in the fine-grained temporal dynamics between speech, gesture, and gaze. A misalignment of even tens of milliseconds can corrupt the analysis of turn-taking, backchannels, or co-speech gestures, and render data unusable for training or evaluating modern audio-conditioned generative models.

Therefore, the design of our capture system treats both forms of synchronization not as ancillary features, but as first-class engineering requirements. Hardware choices, system architecture, and operational procedures are designed to achieve and verify this dual synchronization, producing temporally coherent multi-view, multi-modal recordings suitable for high-fidelity analysis and data-driven modeling.

\subsection{Multi-camera synchronization.} 
\begin{figure}[t]
    \centering
    \includegraphics[width=0.75\linewidth]{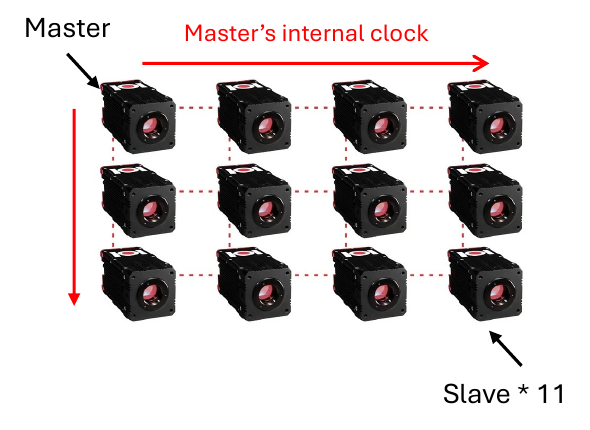}
    \caption{PTP synchronization among cameras. One master camera distributes its internal clock to the remaining slave cameras. Thus, all cameras have a unified internal clock.}
    \label{fig:ptp}
\end{figure}

All 12 machine-vision cameras are connected to the same dedicated 10GbE LAN via the switch, which enables device-level synchronization using the mechanisms supported by the camera platform. In our setup, we employ IEEE-1588 Precision Time Protocol (PTP), which the Volucam cameras support directly over their 10GbE interface. PTP establishes a shared, precise time reference across the array and enables shutter-synchronized multi-camera operation, ensuring that concurrently captured frames are correlated with consistent timestamps. The system is configured in a master–slave topology: one camera serves as the time master, and distributes its clock to the remaining cameras, which operate as slaves and discipline their local clocks accordingly, shown in Fig.~\ref{fig:ptp}. Compared to conventional software-triggered capture (e.g., host-side start commands or per-camera polling), this approach shifts synchronization into the cameras’ timing hardware and avoids operating-system and network scheduling jitter, yielding substantially tighter cross-view alignment for fast motion. For downstream processing, synchronized timestamps are also used as the primary key for associating frames across views. In our deployment, exported image sequences can be named using the master clock time (date + time), and we observed extremely small residual timing differences across cameras; the maximum inter-camera timestamp offset reported during capture was on the order of a few nanoseconds (up to ~6 $ns$) in the best-case configuration.

\subsection{Multi-channel audio synchronization.} While PTP ensures precise temporal alignment across cameras, a complete interaction-capture pipeline also requires that the audio stream be internally consistent and, where applicable, anchored to the same timing reference. Specifically, when recording multiple microphone channels, all channels must be sampled against a single, subtle clock deviations can accumulate into inter-channel drift, degrading downstream processing (e.g., diarization, source separation) and compromising the accuracy of timecode-based audio-video alignment. We therefore begin by describing the synchronization of the multi-channel audio recording chain itself.

The fundamental timing reference for multi-channel audio is the word clock, which determines the precise sampling instants for the analog-to-digital converters (ADCs). A common word clock ensures that samples across all channels correspond to the same moments in time. Our system supports two operational modes to establish this reference: (i) Internal clocking: the audio interface uses its own high-precision internal word clock as the master sampling reference for all channels. (ii) External clocking: an external timecode generator supplies a word clock signal to the audio interface, which then locks its sampling clock to this external reference and distributes it across all channels.

\begin{figure*}[t]
    \centering
    \includegraphics[width=0.75\linewidth]{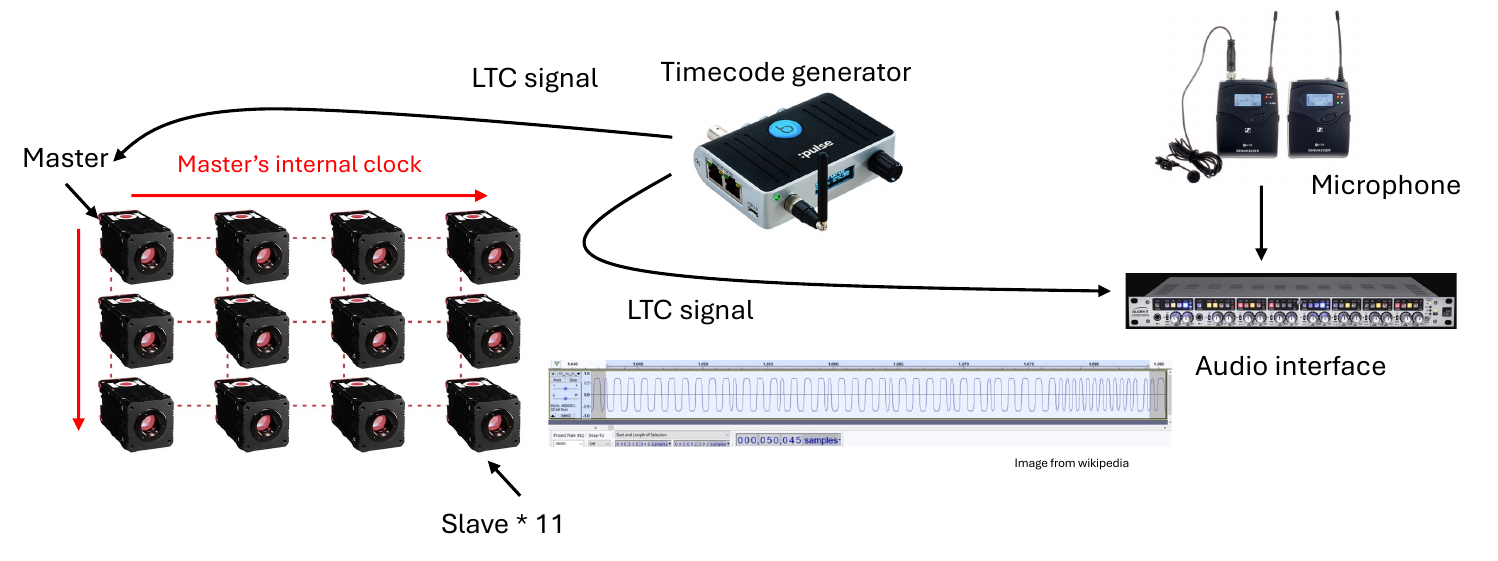}
    \caption{Audio-video synchronization scheme. The tiemcode signal from the timecode generator is split and simultaneously fed to the master camera and, as an audio track, to the audio interface alongside all microphone channels.}
    \label{fig:audio-video sync}
\end{figure*}
\subsection{Audio-video synchronization.} Unlike the PTP-enabled cameras, the audio recording devices (e.g., microphones and the audio interface) do not possess an internal clock that can be aligned with the video through PTP in the same direct manner as the cameras are aligned with each other.

To bridge this gap and establish a unified time reference between the audio and video subsystems, we employ a Linear Timecode (LTC) signal generated by a dedicated master timecode generator. The continuous LTC output is recorded in two parallel paths: one is fed into the synchronization input of the PTP master camera, embedding the timecode into the video metadata; the other is recorded as a separate audio track on the multi-channel audio interface, alongside all microphone channels. This dual recording creates an explicit temporal link: the same timecode exists both in the video stream and as an audio signal.

During post-processing, alignment is achieved by matching the LTC recorded in the audio track with the LTC embedded in the video stream from the master camera. This method maps the audio samples onto the PTP-synchronized camera timeline, enabling consistent audio-video alignment for downstream multimodal analysis. An overview of this synchronization scheme is shown in Fig.~\ref{fig:audio-video sync}.

To establish audio-video alignment in practice, we follow a straightforward, timecode-based procedure, as implemented in our workflow:
\begin{itemize}
    \item System connection: All devices are connected according to the synchronization scheme shown in Fig.~\ref{fig:audio-video sync}. This ensures the timecode generator delivers a continuous Linear Timecode (LTC) signal to the audio recording chain and the master camera.
    \item Recording with Padding: Recording is started and stopped sequentially for audio and video, with the audio recording interval intentionally set to fully encompass the video recording interval. This practice includes extra audio padding at both the beginning and end of the session.
    \item File Acquisition: The output of each session is a multi-channel audio file and multiple video files. This audio file contains the primary microphone channels alongside a dedicated channel that records the raw LTC signal.
    \item Timecode Decoding: The LTC track is decoded using a software LTC decoder to extract the absolute timecode at the start of the recording, expressed in the format HH:MM:SS:FF (where FF denotes the frame index at the configured frame rate).
    \item Timeline Alignment: Using standard video-editing software (e.g., DaVinci Resolve), the audio and video timelines are aligned. This is achieved by trimming the leading and trailing portions of the audio (based on the decoded timecode) so that its start time matches the video's start time. The result is a synchronized audio-video stream ready for downstream processing.
\end{itemize}
This method leverages the fact that trimming audio is a lossless and computationally trivial operation, unlike video trimming which often requires re-encoding. The intentional audio padding provides a flexible buffer to accommodate the sequential start/stop operations, ensuring the complete video interval is covered.

\subsection{Evaluation of audio-video synchronization.} Device-level mechanisms ensure internal synchronization within each subsystem: the cameras share a common PTP master clock, and all audio channels adhere to a unified word-clock reference. Consequently, we do not separately evaluate these intra-subsystem alignments beyond basic functional verification.

However, audio-video synchronization spans two heterogeneous domains, requiring an explicit mapping between their respective time bases. This cross-domain alignment is the most critical and potentially error-prone component of the timing stack, as offsets can be introduced by signal propagation, interface buffering, encoding delays, or timestamp interpretation. To evaluate the performance of this synchronization pipeline, we perform a dedicated calibration measurement. This benchmark quantifies the end-to-end alignment accuracy, confirming that the system can reliably associate speech events with the correct video frames. 

\begin{figure}
    \centering
    \includegraphics[width=0.95\linewidth]{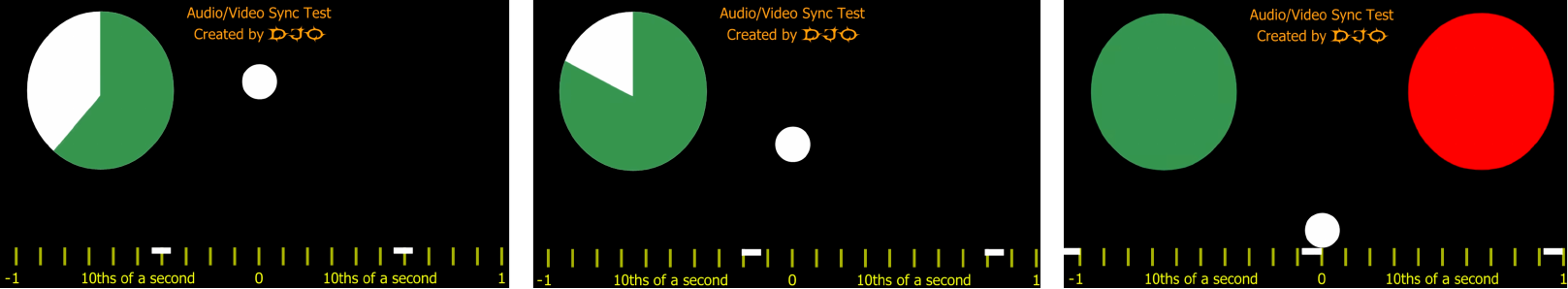}
    \caption{Illustration of the audio-video synchronization test stimulus. The white ball in the middle of the frame drops, and a short white bar moves along the timeline. When they contact, a sound will be made, a red circle will appear in the right part of the frame and the left circle will fully turn to green or white.}
    \label{fig:av test}
\end{figure}
To evaluate the end-to-end audio-visual synchronization accuracy of our pipeline, we employed an event-based method using a publicly available synchronization test stimulus\footnote{https://youtu.be/ucZl6vQ\_8Uo?si=ofLAahfu1hSUSUkO}. This stimulus features a clear, unambiguous audiovisual event: a white ball drops and strikes a horizontal bar, producing a distinct impact sound at the moment of contact, shown in Fig.~\ref{fig:av test}.

\begin{figure}
    \centering
    \includegraphics[width=0.95\linewidth]{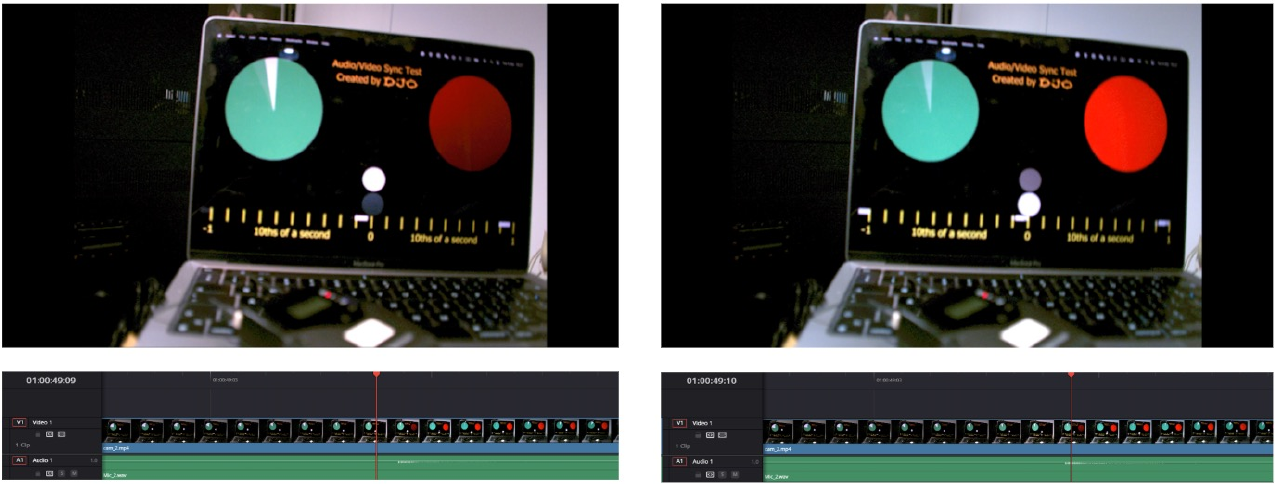}
    \caption{Illustration of the audio-video alignment result. The onset of the impact sound in the audio waveform precedes the visual contact frame by less than one frame ($<$ 1/60 $s$).}
    \label{fig:my sync}
\end{figure}

We recorded this reference clip through our complete capture and alignment workflow. The original stimulus is at 30 fps, while our recording operates at 60 fps. After processing, we imported the synchronized output into a video editing tool to perform frame-accurate and sample-accurate inspection. We manually identified two key points:
\begin{itemize}
    \item Visual contact frame: The first video frame in which the ball makes contact with the bar.
    \item Acoustic onset: The precise sample corresponding to the onset of the impact sound in the audio waveform.
\end{itemize}
In an ideally synchronized system, the audio onset should align temporally with the visual contact frame. In our measurements, the audio onset consistently occurred one video frame earlier than the visual contact frame. This indicates a consistent audio lead of less than one frame duration, which corresponds to $<$ 16.7 ms under our 60 fps recording configuration, shown in Fig.~\ref{fig:my sync}.
\section{Data Acquisition Protocol}
\label{sec:protocol}
The preceding sections have detailed the hardware architecture and synchronization mechanisms of our capture system, which establishes a stable temporal foundation for multi-view audio-visual capture. In practice, however, achieving consistent, high-quality data also relies critically on standardized operational procedures executed before and during each recording session. Variables such as camera white balance, exposure settings, and camera calibration directly influence the repeatability of scene appearance and the accuracy of downstream 3D reconstruction, even when temporal synchronization is perfect. To address these factors and ensure that recordings remain comparable across sessions, we formalize a data acquisition protocol. The protocol prescribes a series of pre-session checks and setup steps, completing the system description with the necessary workflow for reliable and reproducible data collection.

\subsection{Camera white balance.} We perform a manual white balance before each recording session to ensure color consistency across all cameras and over time. While consumer cameras (e.g., smartphones, DSLRs) typically employ automatic white balance (AWB) that continuously adapts to scene content, machine-vision cameras are often configured for predictable, repeatable output. Relying on independent, per-camera AWB in a multi-view setup can introduce view-dependent color shifts, complicating both visualization and downstream appearance-based processing.

\subsection{Procedure.} Our white balance uses a neutral reference target and the cameras' built-in one‑shot white‑balance function:
\begin{itemize}
    \item A matte white sphere is placed within the capture volume, illuminated by the session's lighting and made visible to all cameras.
    \item For each camera, the in-camera white‑balance utility is activated, and a region on the sphere is selected as the neutral reference.
    \item The camera automatically computes the corresponding red‑ and blue‑channel gains from this reference.
    \item These calculated gains are applied and kept fixed for the duration of the session, ensuring consistent color rendering across all twelve views.
\end{itemize}
\subsection{Exposure time and gain.}In addition to white balance, we configure exposure time and sensor gain before each session to obtain images that are both visually usable and suitable for downstream motion analysis. These parameters control a key trade-off: longer exposure admits more light and yields a brighter image, but also increases motion blur during fast head or hand movements; shorter exposure reduces blur but results in a darker image, often requiring an increase in gain to compensate, which in turn amplifies sensor noise. 

In our standard 60 fps recording configuration, we set the exposure time to 3399 $\mu s$. The value was empirically determined to provide a practical balance, sufficiently limiting motion blur while maintaining acceptable image brightness. To achieve the desired brightness under the fixed studio lighting, we set the camera gain to 12.  We treat gain as a secondary control: after selecting an exposure time that satisfies the motion‑blur requirement, gain is adjusted to compensate for the overall illumination, with the understanding that higher gain also scales the sensor’s noise floor.

\subsection{Calibration} Accurate multi-view capture requires precise geometric calibration to obtain the intrinsic and extrinsic camera parameters that define the mapping from 3D world points to 2D image coordinates. These parameters are fundamental for downstream geometry‑based tasks, including image rectification, multi‑view triangulation, and 3D reconstruction.

We perform a joint calibration of the entire camera array using a ChArUco calibration board and the commercial software Calib.io. The procedure involves capturing a sequence of images while moving the calibration board to various positions and orientations throughout the capture volume, ensuring it is visible across the full field of view of all cameras. Calib.io automatically detects the ArUco and checkerboard features on the board and jointly optimizes a consistent set of intrinsic (focal length, principal point, distortion coefficients) and extrinsic (rotation, translation) parameters for all cameras. The resulting calibration is exported in an OpenCV‑compatible format and integrated into our processing pipeline for image undistortion, rectification, and 3D reconstruction from multi‑view detections.
\section{Example datasets}
\label{sec:example_dataset}
The preceding sections have detailed the hardware design, synchronization framework, and operational protocol of our capture system. To demonstrate the practical utility of the data it produces and to validate that the integrated pipeline supports realistic research workflows, we now present two representative data collections acquired with the system. These examples are chosen to illustrate two common and demanding use cases: multi‑person conversational interaction, which challenges the system's multi‑view coverage and precise audio‑visual timing, and single‑subject capture, which demands stable, high‑quality facial and head‑motion signals for training generative models.

\subsection{Multi-person conversational interaction dataset.} To accommodate the spatial requirements of multi‑person interaction, the capture volume is designed to be large enough for 3–6 participants to move and converse naturally. Correspondingly, we equipped the camera array primarily with wide‑field lenses (8$mm$ and 12$mm$) to ensure full coverage of the scene from all views. Leveraging this setup, we captured a dataset comprising approximately 9 hours of synchronized multi‑view video and multi‑channel audio. The dataset consists of 30 conversational sessions across 10 distinct groups, with each group engaging in three different interaction activities: (i) film-inspired discussion, where participants watch a short clip and then converse freely about it; (ii) topic-based debate, where they discuss a given proposition without formal sides; (iii) turn-based games, where participants ask simple questions to deduce the animal named on their own chest. Sessions range from 8 to 29 minutes, with natural variations in group size (3–6 participants) and conversational dynamics.

The recordings are designed to elicit spontaneous multi‑person behavior suitable for downstream modeling of non‑verbal communication, such as co‑speech gestures and turn‑taking, in naturalistic group settings. To balance ecological validity with experimental control, participants were arranged in stable spatial configurations (e.g., seated in a circle).

\begin{figure}[b]
    \centering
    \includegraphics[width=0.95\linewidth]{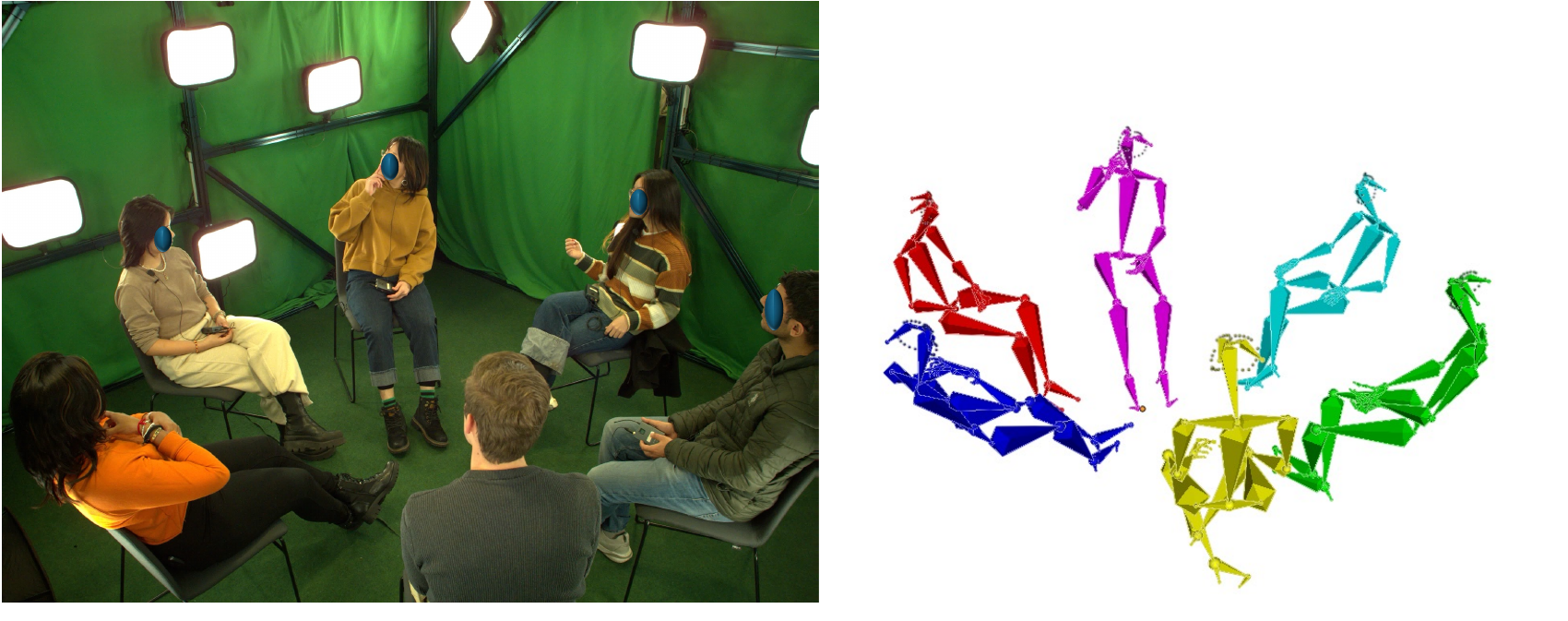}
    \caption{Example from the multi-person conversational interaction dataset. (Left) A raw frame from a single camera view, showing six participants in discussion. To protect participant privacy, all facial visuals presented here have been anonymized through facial masks. The underlying motion and audio data retain their original fidelity and are used for data preprocessing and model training. (Right) The corresponding 3D skeleton keypoints reconstructed from the synchronized multi-view system, demonstrating robust pose estimation despite occlusions.}
    \label{fig:multi-person}
\end{figure}

\begin{figure*}[t]
    \centering
    \includegraphics[width=0.95\linewidth]{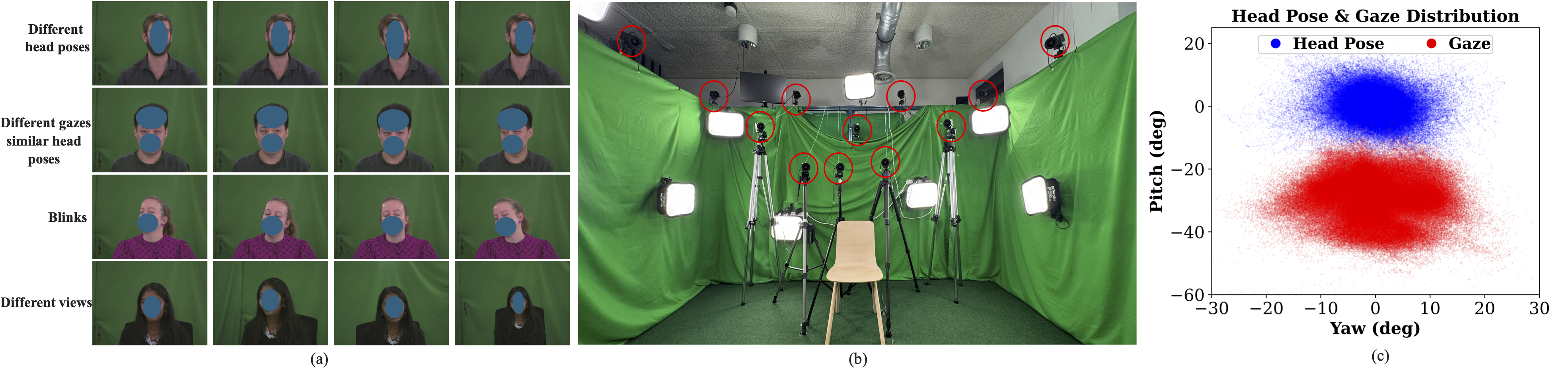}
    \caption{(Left) Example images of single-subject talking head generation dataset, demonstrating different features of the dataset. To protect participant privacy, all facial visuals presented here have been anonymized through facial masks. The underlying motion and audio data retain their original fidelity and are used for data preprocessing and model training. (Middle) Multi-view camera layout optimized for facial capture. Cameras are arranged on tripods to focus on the seated participant, ensuring high-resolution coverage of the face and upper torso from multiple perspectives. (Right) Scatter plot demonstrating the large, natural variation in head pose and gaze direction. Head pose is in the main camera coordinate system, and gaze direction is in the head coordinate system: +x (right), +z (forward toward nose), +y (downward, completing right-handed frame). }
    \label{fig:single}
\end{figure*}

An example of the resulting data is shown in Fig.~\ref{fig:multi-person}. We process the synchronized multi-view video as follows. First, 2D keypoints (body, hands, face) are detected independently in every camera view using OpenPose~\cite{8765346,simon2017hand,cao2017realtime,wei2016cpm}. These 2D detections are then lifted to 3D using the multi-camera calibration parameters within the EasyMocap framework~\cite{shuai2022multinb}. Finally, to obtain temporally smooth 3D trajectories suitable for learning-based modeling, we apply a lightweight Savitzky-Golay filter~\cite{savitzky1964smoothing} to the reconstructed keypoint sequences. The resulting dataset provides spatially and temporally coherent 3D pose estimates, accompanied by precisely aligned multi-channel audio, forming a ready-to-use resource for analyzing and modeling multi-party interaction.

Compared to existing datasets for 3D non‑verbal gesture generation, our collection provides three key advancements: (i) comprehensive modality coverage (full-body, hands, face, and high-fidelity audio), (ii) variable group sizes enabling the study of multi‑party dynamics, and (iii) diverse, structured activities that elicit a wide range of interactive behaviors. The resulting dataset provides spatially and temporally coherent 3D pose estimates, accompanied by precisely aligned multi‑channel audio, forming a ready‑to‑use resource for analyzing and modeling complex social interaction. The aforementioned advantages, such as comprehensive modality coverage, support for variable groups, and activity diversity, are direct outcomes of the capture system’s design, which prioritizes high‑fidelity, synchronized multi‑view audio‑visual recording in a scalable interaction volume.

\subsection{Single-subject talking head generation dataset.} As a second example, we present a single‑subject talking head dataset designed for modeling natural head pose, gaze, and facial dynamics in audio‑driven talking‑head generation. In contrast to the multi‑person scenario, this use case centers on a small region of interest—the participant’s face and upper torso. We therefore reconfigure the capture setup to prioritize facial detail: the \textsc{Sigma} 18–35$mm$ zoom lens is used on all cameras, and a subset of cameras are repositioned onto tripods so that every view is directed toward the participant, yielding a mix of predominantly frontal views supplemented by a small number of side‑profile angles.

To preprocess and annotate the talking‑head data, we follow a pipeline that transforms raw multi‑view recordings into structured representations suitable for generative modeling. First, we segment the audio to remove the facilitator's speech and isolate the participant's speaking intervals using speaker diarization. Then, for each resulting clip and camera view, we extract 2D facial landmarks and per‑frame gaze estimates using OpenFace. These multi‑view 2D landmarks are subsequently triangulated into a robust set of 3D facial landmarks using EasyMocap and the calibrated camera parameters. From the reconstructed 3D landmarks, we estimate the 3D head pose as a rigid transformation relative to a designated main‑camera coordinate system. Finally, per‑view 2D gaze vectors are transformed into the common 3D coordinate system, aggregated across valid views, and expressed as a single 3D gaze vector in the head‑coordinate frame for each time step.

The output of this pipeline is a temporally aligned set of 3D head poses and 3D gaze vectors, accompanied by the corresponding cleaned audio, forming a compact, ready‑to‑use representation for audio‑driven talking‑head generation.

\section{Discussion}
\label{sec:discussion}
The key lesson is that synchronization is the primary design constraint: accurate audio-video alignment is essential for speech-motion timing, and tight multi-camera synchronization is essential for coherent multi-view geometry. Crucially, this also requires stable device-level timing within each camera (trigger response and exposure consistency), otherwise "synchronized" frames may not correspond to the same physical instant. Under a limited budget, we would prioritize camera that support robust hardware synchronization interfaces (e.g., external trigger and genlock/clock I/O). With additional budget, we would first increase camera count to densify viewpoint coverage, which reduces occlusions and typically improves multi-view 3D reconstruction quality; addition viewpoints can also make calibration and reconstruction more robust to occasional tracking failures.
\section{Conclusion}
\label{sec:conclusion}
In this technical report, we have presented a synchronized audio-visual multi-view capture system designed to support research on human interaction where precise timing between speech and motion is essential. Departing from camera‑centric setups that treat audio as secondary, our system integrates multi‑channel audio as a first‑class, temporally aligned modality, enabling recordings that are directly suitable for fine‑grained interaction analysis and modern audio‑conditioned generative modeling.

We have detailed the system design, synchronization mechanisms, and operational protocol that make synchronized capture both practical and repeatable. Empirical validation confirms that our workflow achieves end‑to‑end audio‑video alignment with sub‑frame accuracy. We hope this documentation provides a practical blueprint for building and operating integrated audio‑visual capture infrastructure and, in turn, helps facilitate more reliable, reproducible, and temporally coherent multimodal datasets for future research.

\section*{Acknowledgments}
\noindent We are grateful to Dimme de Groot for his assistance in building the capture system.

{\small
\bibliographystyle{IEEEtran}
\bibliography{references}
\balance
}

\end{document}